\documentclass{article} 
\usepackage{iclr2027_conference,times}

\usepackage{hyperref}
\usepackage{url}

\usepackage[utf8]{inputenc}
\usepackage[T1]{fontenc}
\usepackage{amsmath}
\usepackage{amsfonts}
\usepackage{amssymb}
\usepackage{booktabs}
\usepackage{multirow}
\usepackage{graphicx}
\usepackage{algorithm}
\usepackage{algorithmic}
\usepackage{microtype}
\usepackage{nicefrac}
\usepackage{siunitx}
\usepackage{tikz}
\usepackage{makecell}
\usepackage{wrapfig}
\usepackage[caption=false,font=footnotesize]{subfig}
\usepackage[table]{xcolor}
\usepackage{tcolorbox}
\usepackage{cleveref}
\newtcolorbox[auto counter]{promptbox}[2][]{
  colback=black!3,
  colframe=black!45,
  boxrule=0.4pt,
  arc=0pt,
  left=8pt,
  right=8pt,
  top=7pt,
  bottom=7pt,
  before skip=4pt,
  after skip=8pt,
  fontupper=\small\ttfamily,
  float=htbp,
  title=Prompt~\thetcbcounter: #2,   
  #1
}
\newcommand{\scoreopen}{\tikz[baseline=-0.55ex]{\draw[line width=0.35pt] (0,0) circle (0.75ex);}}
\newcommand{\scorehalf}{\tikz[baseline=-0.55ex]{\begin{scope}\clip (0,0) circle (0.75ex);\fill (0,-1ex) rectangle (1ex,1ex);\end{scope}\draw[line width=0.35pt] (0,0) circle (0.75ex);}}
\newcommand{\scorefull}{\tikz[baseline=-0.55ex]{\filldraw[line width=0.35pt] (0,0) circle (0.75ex);}}

\iclrfinalcopy

\title{TRACE: Trajectory-Based Safety Patch\\Learning for LLM Post-Training Realignment}

\author{Changyue Li$^{1}$\,Jiaming He$^{2}$\,Youliang Yuan$^{3}$\,Jialin Wu$^{4}$\,Boxi Yu$^{5}$\,Zhicong Huang$^{4}\footnotemark[1]$\,\;Pinjia He$^{1}$\thanks{Corresponding authors. This work was completed during Changyue Li's internship at Ant Group. 
}\\
{$^{1}$The Chinese University of Hong Kong, Shenzhen} 
{$^{2}$Nanyang Technological University} \\
{$^{3}$Meituan} 
{$^{4}$Ant Group} 
{$^{5}$Lero the Research Ireland Centre for Software, University of Limerick} \\
\texttt{changyueli@link.cuhk.edu.cn}
}

\begin{document}

\maketitle
\fancyhead{}
\lhead{Preprint}

\begin{abstract}
Fine-Tuning-as-a-Service (FTaaS) platforms let users perform supervised fine-tuning (SFT) on customized data, but this pipeline can erode model safety alignment. To recover safety without re-running full alignment, existing realignment methods focus on calibrating the integration of safety patches into fine-tuned models. These methods exhibit a persistent safety-utility trade-off: weak repair leaves harmful behavior intact, while stronger repair increasingly damages the benign task. This paper shifts the focus from online calibration to offline patch learning and aims to learn a safety patch that restores safety while preserving task-specific capabilities. To this end, we propose \textbf{TRACE}, which simulates harmful SFT trajectories to produce progressively corrupted model states, and optimizes a safety patch simultaneously across these states. TRACE trains the safety patch during the offline stage, and reuses it across all user checkpoints without per-user calibration. We evaluate two representative models using three harmful SFT datasets, together with three utility benchmarks. Across six benchmarks and two models, TRACE consistently dominates the safety--utility frontier. TRACE improves the safety rate by up to 77 percentage points over the best baselines, while maintaining comparable utility to the undefended fine-tuned model.
\end{abstract}

\section{Introduction}
\label{sec:intro}

Commercial Fine-Tuning-as-a-Service (FTaaS) platforms~\citep{openai_finetuning,amazon_bedrock_customization} allow users to upload personal data to customize Large Language Models (LLMs)~\citep{yang2025qwen3,grattafiori2024llama}, where service providers execute the training and deliver the fine-tuned models via APIs. This paradigm lowers infrastructure barriers and protects proprietary weights.

However, FTaaS can also erode the inherent safety alignment of LLMs~\citep{yang2023shadow}, particularly when malicious users fine-tune models on harmful response data. Service providers need a post-training defense that restores safety without re-running full alignment or destroying the utility gained from customized tasks. 

Current realignment research can be categorized into three classes based on the intervention stage, as summarized in \Cref{tab:related_timeline}.
\emph{(i) Pre-FT alignment hardening} methods~\citep{huang2024vaccine,huang2025booster} boost the base model to make safety more resistant to later fine-tuning. 
\emph{(ii) In-FT safety-preserving} methods~\citep{zhang2026spf,li2025salora} constrain the fine-tuning dynamics to retain safety during the training process. However, although both types of methods provide some resistance, the model can still converge to an unsafe state when tuned on enough harmful data.

\emph{(iii) Post-FT merging-based} methods~\citep{bhardwaj2024resta,wu2025enchtable,hsu2024safelora,lu2025safedelta} have emerged to restore safety by integrating a safety patch. These methods focus on calibrating this integration through a merging operator and its repair strength. Although lightweight and efficient, they exhibit a persistent \emph{safety-utility trade-off}. Weak repair may leave harmful behavior intact, whereas stronger repair can degrade task utility~\citep{ilharco2023editing}. Therefore, current merging-based methods fail to simultaneously satisfy the needs of multiple users well.

\begin{table*}[t]
    \centering
    \caption{Comparison of intervention paradigms for post-training safety recovery.}
    \label{tab:related_timeline}
    \footnotesize
    \setlength{\tabcolsep}{5pt}
    \renewcommand{\arraystretch}{1.08}
    \resizebox{\textwidth}{!}{
    \begin{tabular}{cll|ccccc}
        \toprule
        \multicolumn{2}{c}{Paradigm} & \multirow{2}{*}{Methods} & \multirow{2}{*}{Safety} & \multirow{2}{*}{Utility} & \multirow{2}{*}{Robustness} & \multirow{2}{*}{Efficiency} & \multirow{2}{*}{Extensibility} \\
        \cmidrule(lr){1-2}
        Stage & Mechanism & & & & & & \\
        \midrule
        Pre-FT & Alignment Hardening & \shortstack[l]{Vaccine~\citep{huang2024vaccine}, Booster~\citep{huang2025booster}} & \scoreopen & \scorefull & \scoreopen & $0$ & \scorehalf \\
        \midrule
        \addlinespace[2pt]
        In-FT & Safety-preserving FT & \shortstack[l]{SaLoRA~\citep{li2025salora}, SPF~\citep{zhang2026spf}} & \scoreopen & \scorefull & \scorehalf & ${\sim}10^3$ & \scorehalf \\
        \midrule
        \addlinespace[2pt]
        \multirow{5}{*}{Post-FT} & Safety Re-training & OneShot~\citep{zhang2026safety} & \scorehalf & \scorehalf & \scoreopen & ${\sim}10^1$ & \scorehalf \\
        \cmidrule(lr){2-8}
         & Online Merge Calibration & \makecell[l]{RESTA~\citep{bhardwaj2024resta}, EnchTable~\citep{wu2025enchtable}\\SafeLoRA~\citep{hsu2024safelora}, SafeDelta~\citep{lu2025safedelta}} & \scorehalf & \scorehalf & \scoreopen & $10^1 \sim 10^2$ & \scorehalf \\
        \cmidrule(lr){2-8}
         &Offline Patch Learning & \textbf{TRACE (Ours)} & \scorefull & \scorefull & \scorefull & $\sim0.5$ & \scorefull \\
        \bottomrule
    \end{tabular}
    }
    \vspace{2pt}

    \parbox{0.98\textwidth}{\footnotesize \textit{Notes.} Open, half-filled, and filled circles are a qualitative synthesis. \textbf{Safety}: the ability to restore or maintain safety after user fine-tuning. \textbf{Utility}: the ability to retain downstream task performance. \textbf{Robustness}: stability across downstream training intensity. \textbf{Efficiency}: The additional computational overhead (seconds) required for online safety recovery per user. \textbf{Extensibility}: the capability to incorporate other realignment strategies.
    }

\end{table*}

This motivates us to shift the focus from online calibration to offline patch optimization. Instead of calibrating the integration of the safety patch against unpredictable users, we ask: \textbf{Can we find an effective safety patch that restores safety with minimal interference with the user's task performance?}

\begin{figure}[t]
    \centering
    \includegraphics[width=\textwidth]{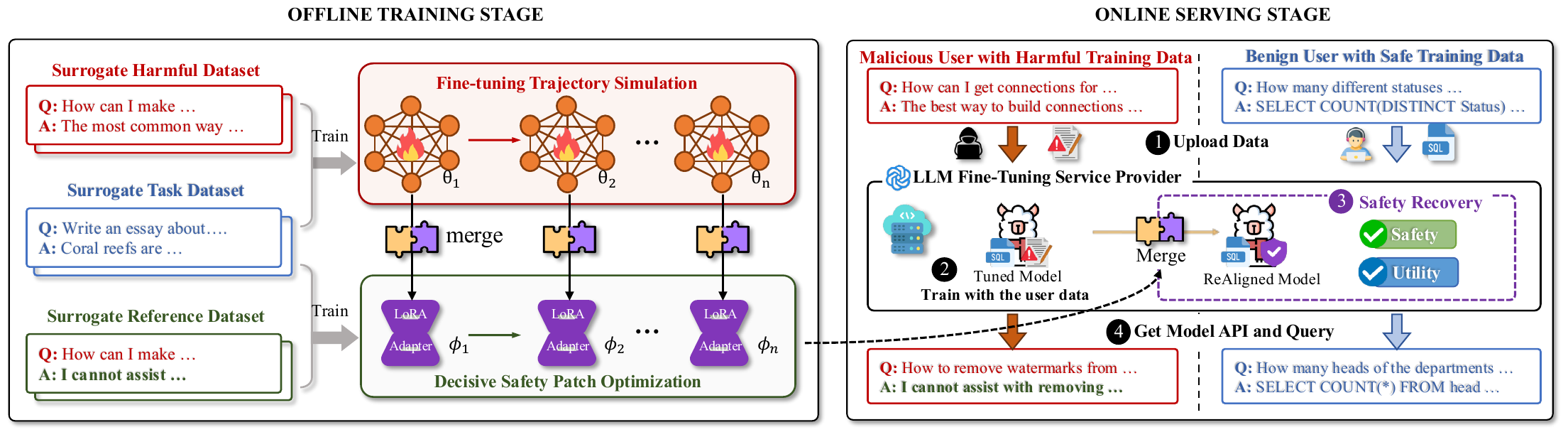}
    \caption{Overview of TRACE. Offline (left), harmful SFT trajectory simulation alternates with safety patch optimization. Online (right), the provider applies the learned patch directly to a fine-tuned model built from the same backbone, without per-user calibration.}
    \label{fig:framework}
\end{figure}

To this end, we propose \textbf{TRACE}, a trajectory-based framework that learns an effective, task-preserving safety patch during the offline stage. As shown in \Cref{fig:framework}, TRACE achieves this through a simulate-and-recover strategy within low-rank space. First, it simulates one fine-tuning trajectory to generate progressively corrupted model states with varying training intensities. Subsequently, TRACE optimizes a low-rank adapter to recover safety across these states while preserving measured task utility. TRACE trains just one patch for each base model. After offline training, the learned patch is directly applied to the user SFT-altered checkpoints built on the same backbone.

Extensive experiments validate the superiority of the proposed method. We evaluate TRACE on two representative models using three harmful datasets~\citep{yang2023shadow,qi2023fine,ji2024pku} of varying sizes alongside three utility datasets spanning dialogue summarization~\citep{gliwa2019samsum}, SQL generation~\citep{b-mc2_2023_sql-create-context}, and mathematical solving~\citep{cobbe2021gsm8k}. The results highlight that TRACE effectively resolves the safety-utility trade-off. Specifically, TRACE improves the safety rate by up to 77 percentage points (pp) over the best baselines, while maintaining downstream task performance within a $\pm$1.7 pp deviation from the undefended baseline.

Our main contributions can be summarized as follows:
\begin{itemize}
    \item We recast post-FT realignment as a safety patch optimization problem, and shift the focus of the merging-based methods from online calibration to offline optimization. 

    \item We propose TRACE, a trajectory-based safety patch learning algorithm. TRACE simulates an SFT trajectory to capture models across varying levels of training intensity, and then optimizes a universal safety patch that consistently restores safety along the trajectory while preserving task utility.
    \item We conduct extensive experiments on two LLMs across three safety and three task datasets. The results demonstrate that TRACE effectively restores safety, achieving over 94\% safety rates while preserving benign task performance.
\end{itemize}

\section{Threat Model and Problem Formulation}
\label{sec:problem}

We consider an FTaaS scenario involving two entities, the service provider and the user. The provider hosts an aligned base model $\theta_{\text{base}}$ and performs full-parameter supervised fine-tuning (SFT) on the user-uploaded data, yielding the fine-tuned model $\theta_{\text{ft}}$. The user controls the dataset and permitted SFT hyperparameters, including epochs and learning rate. The provider prepares a safety patch offline and applies it after user training. Under our deployment assumption, the service exposes the final model through black-box inference APIs and does not release the hosted weights or the patch.  
The primary objective of the service provider is to deliver inference APIs that guarantee both model safety and downstream task utility.


To recover the safety of $\theta_{\text{ft}}$, merging-based methods construct a safety patch $\Delta \theta_{\text{patch}}$ offline and merge it online with the user update. During the offline stage, the patch is typically obtained~\citep{bhardwaj2024resta,wu2025enchtable} by training an intentionally unsafe model $\theta_{\text{unsafe}}$ on surrogate harmful data and taking the parameter difference from the base model (i.e., $\Delta \theta_{\text{patch}} = \theta_{\text{base}} - \theta_{\text{unsafe}}$). During the online stage, the safety repair process can be formulated as:
\begin{equation}
    \label{eq:merging_paradigm}
    \theta_{\text{deploy}} = \theta_{\text{base}} \oplus \Delta \theta_{\text{user}} \oplus \Delta \theta_{\text{patch}}
\end{equation}
where $\Delta \theta_{\text{user}}$ is the parameter update from the user's data and $\oplus$ is the merging operator. \textit{Arithmetic-based} methods~\citep{bhardwaj2024resta,wu2025enchtable} scale the repair strength by a coefficient $\lambda$, $\theta_{\text{deploy}} = \theta_{\text{base}} + \Delta \theta_{\text{user}} + \lambda \Delta \theta_{\text{patch}}$; \textit{projection-based} methods~\citep{hsu2024safelora,lu2025safedelta} map both updates through projection matrices $P_1(\alpha), P_2(\alpha)$ before merging (i.e., $\theta_{\text{deploy}} = \theta_{\text{base}} + P_1(\alpha) \Delta \theta_{\text{user}} + P_2(\alpha) \Delta \theta_{\text{patch}}$),
where $\alpha$ adjusts the repair strength. 

In real-world FTaaS deployments, users can choose the uploaded dataset, number of training epochs, and learning rate~\citep{openai_finetuning,azure_ai_finetuning,amazon_bedrock_customization,google_vertex_tuning}. As a result, different users may adopt substantially different training configurations, which the provider cannot fully anticipate in advance. The static repair strength employed by existing realignment methods may therefore be inadequate to accommodate such diverse user requests. We evaluate the resulting safety--utility trade-off across users in \Cref{sec:safety_utility_across_users}.


This observation motivates a shift from online calibration to offline patch learning. During offline training, the provider observes only surrogate data distributions, with private user distributions remaining inaccessible. The safety repair on user-tuned checkpoints is therefore recast as a deployment-time safety transfer problem. We formalize this safety-patch optimization problem in the next section and introduce TRACE's alternating recovery procedure.
\section{Trajectory-Based Safety Patch Learning}
\label{sec:method}

This paper shifts the focus to offline optimization of a transferable safety patch $\Delta \theta_{\text{patch}}$ for an unseen SFT-altered model $\theta_{\text{ft}}$.
We denote the learned patch as $\phi$ to distinguish it from the conventional safety patch $\Delta\theta_{\text{patch}}$ in \Cref{sec:problem}.

The safety patch should elicit refusal responses on harmful inputs while preserving the model's downstream task performance. Thus, we have the following objective:
\begin{equation}
   \phi^{*} = 
    \arg\min_{\phi} \;
    \mathbb{E}_{(x,y) \sim \{\mathcal{U}_{\text{safe}},\mathcal{U}_{\text{task}}\}}
    \left[\mathcal{L}(\theta_{\text{ft}} \oplus \phi, x, y)\right],
    \label{eq:ideal_patch_objective}
\end{equation}
where $\mathcal{U}_{\text{safe}}$ contains harmful questions paired with safe responses, and $\mathcal{U}_{\text{task}}$ contains task-related questions and their corresponding answers. Once the patch is obtained offline, the provider directly deploys $\theta_{\text{deploy}} = \theta_{\text{ft}} \oplus \phi^{*}$ without online per-user adaptation.

Since the user-uploaded dataset $\mathcal{U}$ is unavailable during the offline training, TRACE instead approximates the objectives via provider-owned surrogate datasets, expecting the resulting safety capability to transfer effectively to the user-customized model. The provider trains just one patch $\phi$ for each base model. For a user fine-tuned model $\theta_{\text{ft}}$ built on the same backbone, the provider can simply apply direct weight merging ($\theta_{\text{ft}} + \phi$) to restore safety alignment, eliminating the need for per-user adjustments or projections. Notably, since TRACE aims to search for a better safety patch, it is highly compatible with existing techniques, seamlessly leveraging their online merging operations to further boost defensive performance.

\subsection{Alternating Safety Recovery Training}
\label{sec:alternating}

Ideally, the safety patch $\phi$ should be optimized over all corrupted states ${\theta_\text{ft}}$ as described in \Cref{eq:ideal_patch_objective}. 
However, directly optimizing this objective is impractical, as it requires caching every possible model state that could arise during fine-tuning. For LLMs, storing and processing such dense intermediate checkpoints incurs prohibitive memory and computational overhead.

TRACE adopts an alternating optimization strategy to approximate this expectation efficiently. At each iteration, TRACE first simulates the user's full-parameter fine-tuning and obtains a freshly updated model state, and then TRACE immediately optimizes the safety patch on top of it to restore safety while preserving task performance. In this way, TRACE approximates optimization over the trajectory without storing every model state. We acknowledge that this alternating procedure only approximates the ideal objective and inevitably introduces some discrepancy. Nevertheless, the primary goal of this work is not to construct the optimal patch, but rather to demonstrate this safety patch learning paradigm provides an effective and practical strategy for model realignment.


\noindent\textbf{Phase A: simulate safety-eroding fine-tuning.}
TRACE simulates user full parameter fine-tuning using provider-owned surrogate harmful data $\mathcal{D}_{\text{harm}}$ and benign data $\mathcal{D}_{\text{task}}$. The trajectory therefore reflects both harmful drift and ordinary task adaptation. For a data batch $B$ and model parameters $\theta$, we use the response-only causal language modeling loss
\begin{equation}
    \mathcal{L}(\theta; B) = -\sum_{x \in B} \sum_{t \in \mathcal{A}(x)} \log p_{\theta}(x_t \mid x_{<t}),
    \label{eq:ce_loss}
\end{equation}
where $\mathcal{A}(x)$ denotes the assistant-response token positions in example $x$.

For the current iteration $t$, we disable the safety patch and update only the backbone model on one harmful batch and one benign task batch:
\begin{equation}
    \theta_t = \theta_{t-1} - \eta_{\theta} \nabla_{\theta}
    \mathcal{L}(\theta_{t-1}; \{B_h, B_t\})
    \label{eq:inner_update}
\end{equation}
where $\eta_{\theta}$ is the learning rate for the backbone model, $\theta_0=\theta_\text{base}$, and $B_h \sim \mathcal{D}_{\text{harm}},\; B_t \sim \mathcal{D}_{\text{task}}$ are data samples. As each update combines harmful and benign supervision, this trajectory captures both safety erosion and ordinary task adaptation at progressively greater fine-tuning intensities.


\noindent\textbf{Phase B: optimize the safety patch.}
Given the newly corrupted state $\theta_t$, TRACE freezes the backbone, enables the safety patch, and updates it using refusal supervision $\mathcal{D}_{\text{ref}}$ and benign task data $\mathcal{D}_{\text{task}}$. The patch update is
\begin{equation}
    \phi_t = \phi_{t-1} - \eta_{\phi} \nabla_{\phi}
    \mathcal{L}\left(\theta_t \oplus \phi_{t-1};  \left\{B_{\text{ref}}, B_t\right\}\right),
    \label{eq:patch_update}
\end{equation}
where $\eta_{\phi}$ is the patch learning rate, $\phi_0$ is zero-initialized, and $B_{\text{ref}}$ are sampled from $\mathcal{D}_{\text{ref}}$. 
This objective trains the patch to recover refusal behavior at the current corruption state $\theta_t$ while preserving responses on the surrogate benign tasks. 

By alternatively repeating the above two phases, TRACE requires the safety patch to consistently succeed across multiple evolving corrupted states rather than overfitting to one fixed checkpoint. Finally, we obtain the learned safety patch $\phi^{*}$ without explicitly storing intermediate checkpoints. 

We distinguish two behavioral goals from possible geometric explanations:

\noindent\textbf{Limited task interference.} The patch optimization objective (\Cref{eq:inner_update}) jointly enforces refusal supervision and benign task preservation across evolving states, which we hypothesize encourages patch updates that interfere minimally with task behavior.

\noindent\textbf{Decisive refusal recovery.} The learned safety patch should reliably restore refusal behavior even in the presence of harmful weight shifts of unknown intensity. We conjecture that this requires the patch to dominate the harmful direction in order to recover safety.

\subsection{Decisive Safety in Low-Rank Subspace}
\label{sec:lowrank}

This paper evaluates the challenging setting where the user performs \emph{full-parameter SFT}, which can induce stronger and more distributed weight drift than parameter-efficient fine-tuning (PEFT).
Consequently, existing realignment methods~\citep{bhardwaj2024resta,lu2025safedelta} typically necessitate full-parameter safety patches. For example, SafeLoRA~\citep{hsu2024safelora} restricts its LoRA-based repair to PEFT models, while relying on full-parameter patches for full-parameter SFT models.

TRACE departs from this design, as we empirically find that even for full-parameter altered models, an effective safety patch can be realized within a low-rank subspace. This highlights an intriguing mechanism wherein, despite extensive shifts in the global weight space, safety capabilities remain anchored along a small number of critical directions~\citep{refusal2024} that exert decisive control over behavior. 
We therefore implement $\phi$ as a LoRA adapter~\citep{hu2022lora}, which significantly reduces trainable parameters during offline optimization and yields a lightweight patch for efficient online deployment.

\subsection{Calibration-Free Deployment and Compatibility}
\label{sec:calibration}

After offline optimization, we freeze the patch and reuse it across user checkpoints derived from the same base model. Under the default deployment, the provider directly applies $\theta_{\text{deploy}} = \theta_{\text{ft}} + \phi^{*}$. This procedure requires neither access to the user-uploaded data nor checkpoint-specific safety probes, gradient updates, or coefficient searches. The offline optimization cost can therefore be amortized across users, while online deployment only requires loading or merging the lightweight adapter. We refer to this setting as \emph{calibration-free} within the target fine-tuning regime.

TRACE is also compatible with prior post-training realignment methods. Existing methods predominantly focus on designing the merging operation $\oplus$, whereas TRACE improves the safety patch $\phi$ itself. Because the patch and the merging rule are orthogonal, the conventional safety patch in an arithmetic- or projection-based method can be replaced by $\phi$ without changing that method's original operator. Such composition is optional and potentially boosts the performance of existing approaches. In \Cref{sec:plugin_composition}, we evaluate this compatibility and measure how it changes the safety-utility curves of several merging methods.

\section{Experiments}
\label{sec:experiments}


\subsection{Experiment Settings}
\label{subsec:setup}

\noindent\textbf{Base models.}
We evaluate on two representative open-source models: \textit{Llama-3.1-8B-Instruct (Llama)}~\citep{grattafiori2024llama}, widely used in prior realignment work~\citep{qi2023fine,yang2023shadow,bhardwaj2024resta}, and \textit{Qwen3.5-9B (Qwen)}~\citep{team2026qwen3}, a more recent model to test TRACE effectiveness across model families. 

\noindent\textbf{Datasets.}
Post-FT realignment involves an \emph{offline} stage, where the provider prepares the safety patch, and an \emph{online} stage, where user checkpoints are evaluated. Offline, following prior work~\citep{wu2025enchtable, yuan2025refuse}, we sample 5{,}000 examples each from \textit{BeaverTails}~\citep{ji2024beavertails} (harmful) and \textit{WizardLM}~\citep{xu2023wizardlm} (benign). Online, \emph{malicious} users train on 100 \textit{ShadowAlignment}~\citep{yang2023shadow}, 500 \textit{PureBad}~\citep{qi2023fine}, or 1{,}000 \textit{SafeRLHF}~\citep{ji2024pku} examples. These represent distinct user corpora with different sizes. \emph{Benign} users train on \textit{SamSum}~\citep{gliwa2019samsum}, \textit{SQLCreate}~\citep{b-mc2_2023_sql-create-context}, or \textit{GSM8K}~\citep{cobbe2021gsm8k}, using the task dataset sizes and splits of OneShot~\citep{zhang2026safety}. \emph{Mixed} corpora combine each task dataset with PureBad. Unless otherwise specified, we train the model on the user data for five epochs.
More implementation details are specified in \Cref{sec:implementation}.


\noindent\textbf{Baselines.}
We evaluate our method against six established baselines. Four are post-FT merging methods closely aligned with our setting: RESTA~\citep{bhardwaj2024resta} and EnchTable~\citep{wu2025enchtable} are arithmetic-based methods that scale and add a safety patch to the tuned LLM, whereas SafeLoRA~\citep{hsu2024safelora} and SafeDelta~\citep{lu2025safedelta} are projection-based methods that suppress safety-sensitive components within the user update. For broader comparison, we include OneShot~\citep{zhang2026safety}, another post-FT approach that performs additional safety fine-tuning on the compromised model to restore alignment, and SPF~\citep{zhang2026spf}, an in-FT realignment method noted for its effectiveness across different training intensities. 

\noindent\textbf{Metrics.}
For safety evaluation, we closely follow prior work~\citep{zhang2026safety} and adopt the held-out benchmark (\textit{StrongReject}~\citep{souly2024strongreject}) to evaluate the \textit{Safety Rate (\%)} on the unseen data distribution. The safety rate is defined as the fraction of model responses that are deemed safe. To judge whether a given response is safe, we use safety-expert discriminators: Llama-Guard-3-8B (LlamaGuard)~\citep{inan2023llama, grattafiori2024llama} for Llama-based models and Qwen3Guard-8B (QwenGuard)~\citep{zhao2025qwen3guard} for Qwen-based models. Since the question-answer pairs are explicitly designed to be harmful, this evaluation constitutes a relatively straightforward task for dedicated safety classifiers. During our experiments, we also manually inspect a subset of the classification results and do not observe any apparent misjudgments. 

For utility metrics, we evaluate each customized task on its respective test set and report \textit{Task Accuracy (\%)}. Since the three datasets differ in output format, we adopt tailored evaluation criteria. Specifically, we leverage Abstract Syntax Tree Matching~\citep{ast} for SQLCreate, which compares the parsed structures of the predicted and reference SQL queries, thereby accommodating syntactically different but semantically equivalent query formulations. We use Exact Match for GSM8K, which checks whether the extracted numerical answer is identical to the ground truth. We report the ROUGE-1 F1 Score~\citep{lin2004rouge} for SamSum, which measures unigram overlap between generated and reference summaries to balance content coverage and conciseness. 

\subsection{Overall Performance}
\label{subsec:main_results}

This section evaluates TRACE under two deployment scenarios, pure-profile fine-tuning and mixed-profile fine-tuning. We analyze safety recovery and utility preservation for each setting. The results are shown in Table~\ref{tab:main_results}, with safety in the upper panel and task accuracy in the lower panel.

\noindent\textbf{Pure profile.}
\label{subsec:pure_profile}
To independently assess safety recovery and utility preservation, we first isolate two extreme cases: malicious users fine-tuning exclusively on harmful data, and benign users fine-tuning on clean task data. We evaluate all methods on both base models across three harmful datasets and three custom task datasets.

\begin{table*}[t]
  \centering
  \caption{Safety rate and task accuracy under pure and mixed profiles.
  In each column, \textbf{bold} and \underline{underline} indicate the best and second-best safety rate, respectively.
  For TRACE, \textcolor{green!60!black}{$\uparrow$} denotes the gain over the second-best value, while \textcolor{blue!70!black}{$\Delta$} denotes the change in task accuracy relative to No Defense.}
  \label{tab:main_results}
  \setlength{\tabcolsep}{3pt}
  \resizebox{\textwidth}{!}{%
  \begin{tabular}{l ccc ccc | ccc ccc}
    \toprule

    \multicolumn{13}{c}{\textbf{Safety Rate (\%) $\uparrow$}} \\
    \midrule
    & \multicolumn{6}{c}{\textbf{Pure Profile}}
    & \multicolumn{6}{c}{\textbf{Mixed Profile}} \\
    \cmidrule(lr){2-7} \cmidrule(lr){8-13}

    & \multicolumn{3}{c}{\textbf{Llama}}
    & \multicolumn{3}{c}{\textbf{Qwen}}
    & \multicolumn{3}{c}{\textbf{Llama}}
    & \multicolumn{3}{c}{\textbf{Qwen}} \\
    \cmidrule(lr){2-4} \cmidrule(lr){5-7}
    \cmidrule(lr){8-10} \cmidrule(lr){11-13}

    \textbf{Dataset}
    & \textbf{Shadow} & \textbf{PureBad} & \textbf{SafeRLHF}
    & \textbf{Shadow} & \textbf{PureBad} & \textbf{SafeRLHF}
    & \textbf{MixedSamSum} & \textbf{MixedSQL} & \textbf{MixedGSM8K}
    & \textbf{MixedSamSum} & \textbf{MixedSQL} & \textbf{MixedGSM8K} \\
    \midrule

    No Defense
      & 18 & 10 & 11
      & 3 & 5 & 2
      & 11 & 20 & 12
      & 2 & 4 & 1 \\

    RESTA
      & \underline{99} & 15 & 8
      & 10 & 3 & 3
      & 13 & 21 & 21
      & 30 & 4 & 50 \\

    EnchTable
      & \textbf{100} & 16 & 10
      & 10 & 21 & 4
      & 14 & 21 & 22
      & 24 & 5 & 44 \\

    SafeLoRA
      & 18 & 8 & \underline{23}
      & 4 & \underline{53} & \underline{38}
      & 83 & \underline{93} & 92
      & 86 & 68 & \underline{99} \\

    SafeDelta
      & 18 & 12 & 15
      & \underline{44} & 15 & 21
      & \underline{98} & \textbf{98} & \underline{98}
      & \textbf{100} & \textbf{100} & \textbf{100} \\

    SPF
      & 11 & 11 & 10
      & 4 & 4 & 0
      & 20 & 20 & 15
      & 3 & 5 & 4 \\

    OneShot
      & \textbf{100} & \underline{67} & 13
      & 36 & \underline{53} & 2
      & 47 & 35 & 33
      & 24 & 7 & 40 \\

    \midrule

    \textbf{TRACE}
      & \textbf{100}
      & \textbf{100}{\scriptsize\textcolor{green!60!black}{$\uparrow$33}}
      & \textbf{100}{\scriptsize\textcolor{green!60!black}{$\uparrow$77}}
      & \textbf{100}{\scriptsize\textcolor{green!60!black}{$\uparrow$56}}
      & \textbf{100}{\scriptsize\textcolor{green!60!black}{$\uparrow$47}}
      & \textbf{100}{\scriptsize\textcolor{green!60!black}{$\uparrow$62}}
      & \textbf{100} & \textbf{98} & \textbf{100}
      & \underline{98} & \underline{94} & \textbf{100} \\

    \midrule\midrule

    \multicolumn{13}{c}{\textbf{Task Accuracy (\%) $\uparrow$}} \\

    \cmidrule(lr){2-4} \cmidrule(lr){5-7}
    \cmidrule(lr){8-10} \cmidrule(lr){11-13}

    & \textbf{SamSum} & \textbf{SQL} & \textbf{GSM8K}
    & \textbf{SamSum} & \textbf{SQL} & \textbf{GSM8K}
    & \textbf{MixedSamSum} & \textbf{MixedSQL} & \textbf{MixedGSM8K}
    & \textbf{MixedSamSum} & \textbf{MixedSQL} & \textbf{MixedGSM8K} \\
    \midrule

    No Defense
      & 51 & 81 & 55
      & 46 & 66 & 42
      & 47 & 76 & 40
      & 50 & 77 & 59 \\

    \textbf{TRACE}
      & 50{\scriptsize\textcolor{blue!70!black}{$\Delta$-0.2}}
      & 81{\scriptsize\textcolor{blue!70!black}{$\Delta$0.0}}
      & 54{\scriptsize\textcolor{blue!70!black}{$\Delta$-0.5}}
      & 46{\scriptsize\textcolor{blue!70!black}{$\Delta$+0.4}}
      & 66{\scriptsize\textcolor{blue!70!black}{$\Delta$+0.1}}
      & 42{\scriptsize\textcolor{blue!70!black}{$\Delta$0.0}}
      & 47{\scriptsize\textcolor{blue!70!black}{$\Delta$-0.6}}
      & 77{\scriptsize\textcolor{blue!70!black}{$\Delta$+1.0}}
      & 40{\scriptsize\textcolor{blue!70!black}{$\Delta$-0.1}}
      & 50{\scriptsize\textcolor{blue!70!black}{$\Delta$-0.1}}
      & 78{\scriptsize\textcolor{blue!70!black}{$\Delta$+0.8}}
      & 57{\scriptsize\textcolor{blue!70!black}{$\Delta$-1.7}} \\

    \bottomrule
  \end{tabular}}%
\end{table*}

As illustrated in \Cref{tab:main_results}, TRACE obtains 100\% safety rates across six pure-profile combinations of benchmarks and models. Its advantage is particularly notable on the challenging SafeRLHF benchmark. TRACE improves the Llama safety rate from 23\% to 100\%, which is over four times higher than the second-best value. Similarly, it increases the Qwen safety rate from 38\% to 100\%, outperforming the second-best baseline by 2.6 times.

TRACE preserves downstream task performance with negligible interference to user tasks. To quantify this, we leverage the \textit{No Defense} approach as a baseline and calculate the utility deviation ($\Delta$) to measure the impact of TRACE on performance. Across the six pure-profile settings, these utility deviations fall within a narrow margin of $\pm$0.5 pp. This near-zero interference demonstrates that the learned safety patch operates along directions largely disentangled from user task updates, preventing it from distorting the learned task behavior. 



\noindent\textbf{Mixed profile.}
\label{subsec:mixed_profile}
This setting evaluates a mixed training scenario where the user corpus contains both benign task examples and harmful content, aiming to test whether TRACE can selectively eliminate harmful behaviors without degrading co-trained task performance. 

TRACE effectively isolates and neutralizes harmful components entangled with task-relevant updates in a mixed training corpus. Across all six mixed configurations, TRACE achieves a safety rate of at least 94\% while keeping task accuracy within $\pm$1.7 pp of No Defense.

Projection-based defenses (e.g., SafeLoRA and SafeDelta) achieve higher safety in the mixed setting but at a significant utility cost. SafeDelta reaches over 98\% safety across all six mixed configurations, showing a sharp increase from its performance on purely malicious data in \Cref{tab:main_results}. This improvement may stem from the larger aggregate model modification induced by jointly training on malicious and benign task data, which causes SafeDelta's projection mechanism to suppress a larger portion of the resulting update. However, this stronger suppression simultaneously destroys co-trained task representations. Full task-accuracy results for all baselines are reported in Appendix~\ref{sec:full_utility_results}, which show that SafeDelta drops MixedGSM8K on Llama from 40\% to 2\%, and SafeLoRA collapses MixedSamSum on Llama from 47\% to 8\%. 

\begin{figure*}[t]
    \centering
    \subfloat[Safety--Utility trade-off curves.\label{fig:combined_alpha}]{%
        \includegraphics[width=0.67\linewidth]{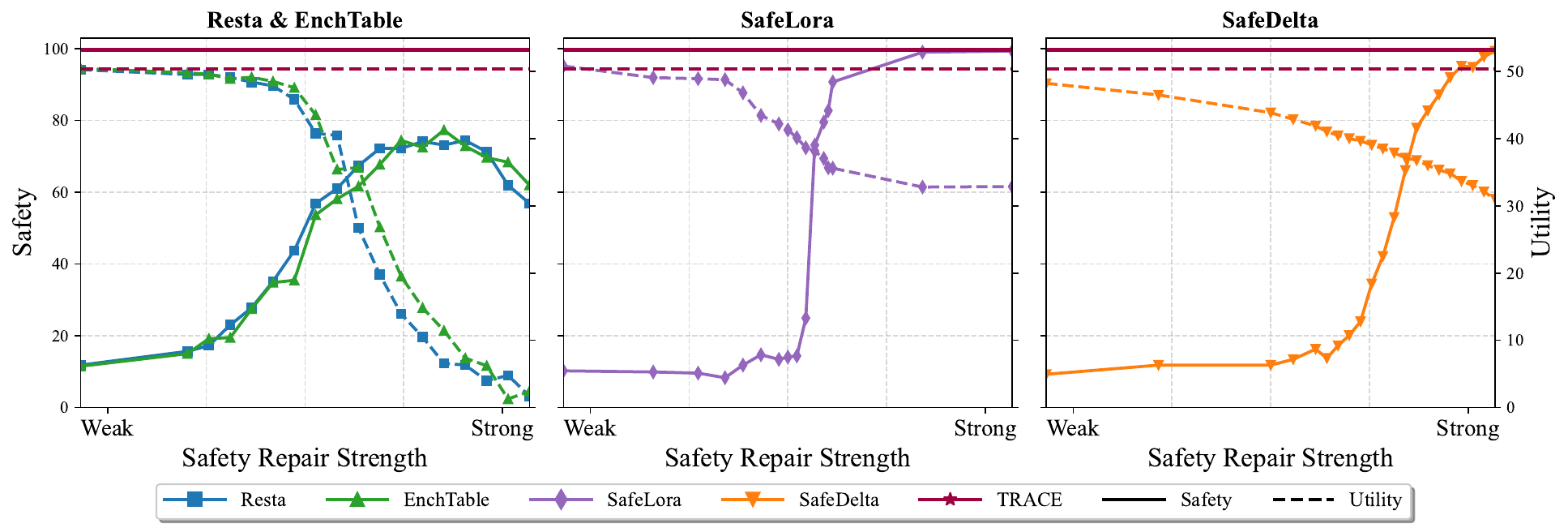}}
     \subfloat[Operating envelope across users.\label{fig:overall_performance}]{%
        \includegraphics[width=0.33\linewidth]{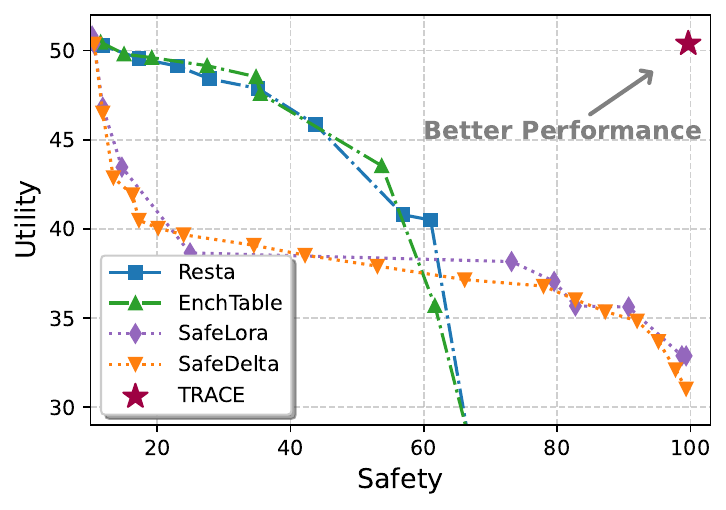}}
    \caption{\textbf{Left:} Safety and utility measured on two representative users as the repair strength varies. \textbf{Right:} Operating envelope across users available to deployed realignment methods. TRACE achieves a favorable operating point without coefficient search.}
    \label{fig:safety_utility_tradeoff}
\end{figure*}

\begin{figure*}[t]
    \centering
    \includegraphics[width=1\linewidth]{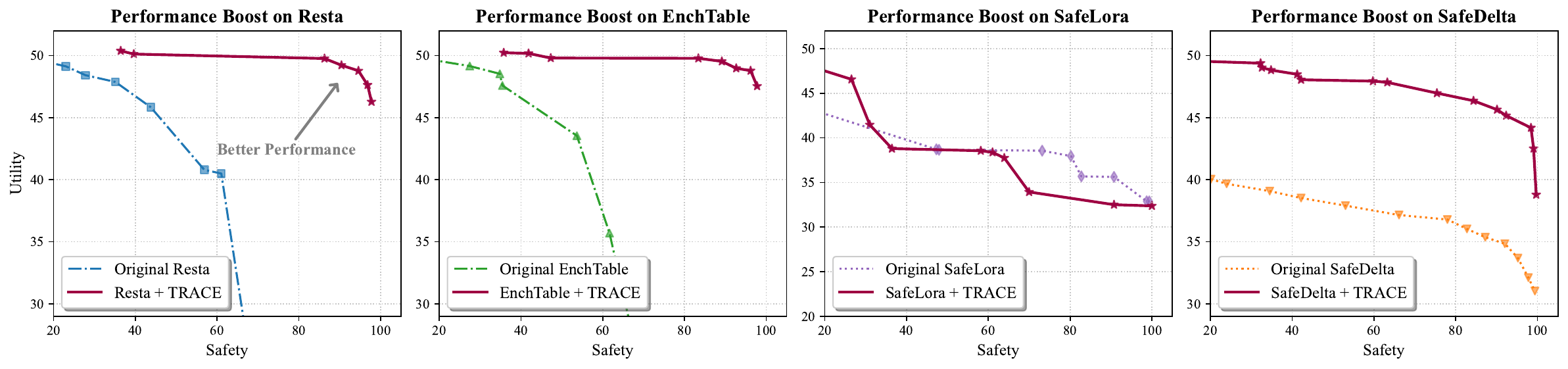}
    \caption{Operating envelopes across users before (dashed) and after (solid) substituting the TRACE patch into each baseline's merging operation.}
    \label{fig:pareto_integration}
\end{figure*}

\subsection{Safety--Utility Trade-off across Users}
\label{sec:safety_utility_across_users}

Although existing merging-based methods~\citep{hsu2024safelora,lu2025safedelta} use adaptive rescaling or projection rules for online calibration, their deployment still depends on a repair-strength coefficient, such as $\lambda$ or $\alpha$, to control the overall intervention magnitude. In an FTaaS deployment, the provider must serve heterogeneous user requests. This raises a deployment issue: a shared coefficient cannot generally provide a favorable operating point for all users.

We evaluate this service-level limitation using two representative user requests: a malicious checkpoint produced from PureBad and a benign checkpoint produced from SamSum. For each baseline, we sweep the repair strength and report the performance obtained using a shared coefficient across both requests. This shared policy reflects the deployment of realignment methods in real-world FTaaS settings. Appendix~\ref{sec:strength_sweep} details the sweep coverage and the method-specific strength axes.

Existing baselines exhibit a clear safety--utility trade-off across users. As shown in \Cref{fig:combined_alpha}, increasing the repair strength improves safety for the malicious user but reduces utility for the benign user. For example, the safety rate of RESTA increases from 10.5\% to 74.4\%, while the task accuracy declines from 50.4\% to 6.3\%. 

TRACE achieves a favorable performance across users without coefficient search. As shown in \Cref{fig:overall_performance}, TRACE reaches a 99.7\% safety rate and 50.4\% task accuracy. It occupies a separate region that the baselines cannot reach with any shared repair strength.


\subsection{Integration and Boost on Existing Methods}
\label{sec:plugin_composition}
Recall from \Cref{sec:calibration} that TRACE optimizes the safety patch $\phi$ itself rather than the merging operation $\oplus$. It is therefore compatible with existing merging-based defenses. This section evaluates whether TRACE can integrate with these baselines and further boost their performance. Specifically, we substitute the standard safety patch in other baselines with the TRACE adapter. We then re-sweep their repair strengths to characterize the resulting operating envelopes across users.

TRACE can be seamlessly integrated with existing baselines and boosts the performance of most methods. As shown in Figure~\ref{fig:pareto_integration}, for RESTA, the integration moves from 74.4\% safety at 6.3\% utility to 97.8\% at 46.3\%. EnchTable and SafeDelta exhibit similar improvements. These results support the value of the learned patch for these merging-based methods.

Unlike other baselines, SafeLoRA gains no explicit benefit from this integration and achieves comparable performance to the original. For instance, at a 60\% safety rate, both the original and integrated versions achieve approximately 38\% utility. A possible explanation is that SafeLoRA uses the patch as a projection reference without directly adding its weights. 

While integrating TRACE improves the safety--utility operating envelopes across users for existing merging strategies, current methods remain constrained because they are designed around conventional patches. For example, the SafeDelta integration still fails to match the standalone performance of TRACE. This likely occurs because SafeDelta alters components of the safety patch to avoid task interference, which may compromise the effectiveness of TRACE's patch.

\begin{figure*}[t]
    \centering
    \includegraphics[width=0.95\linewidth]{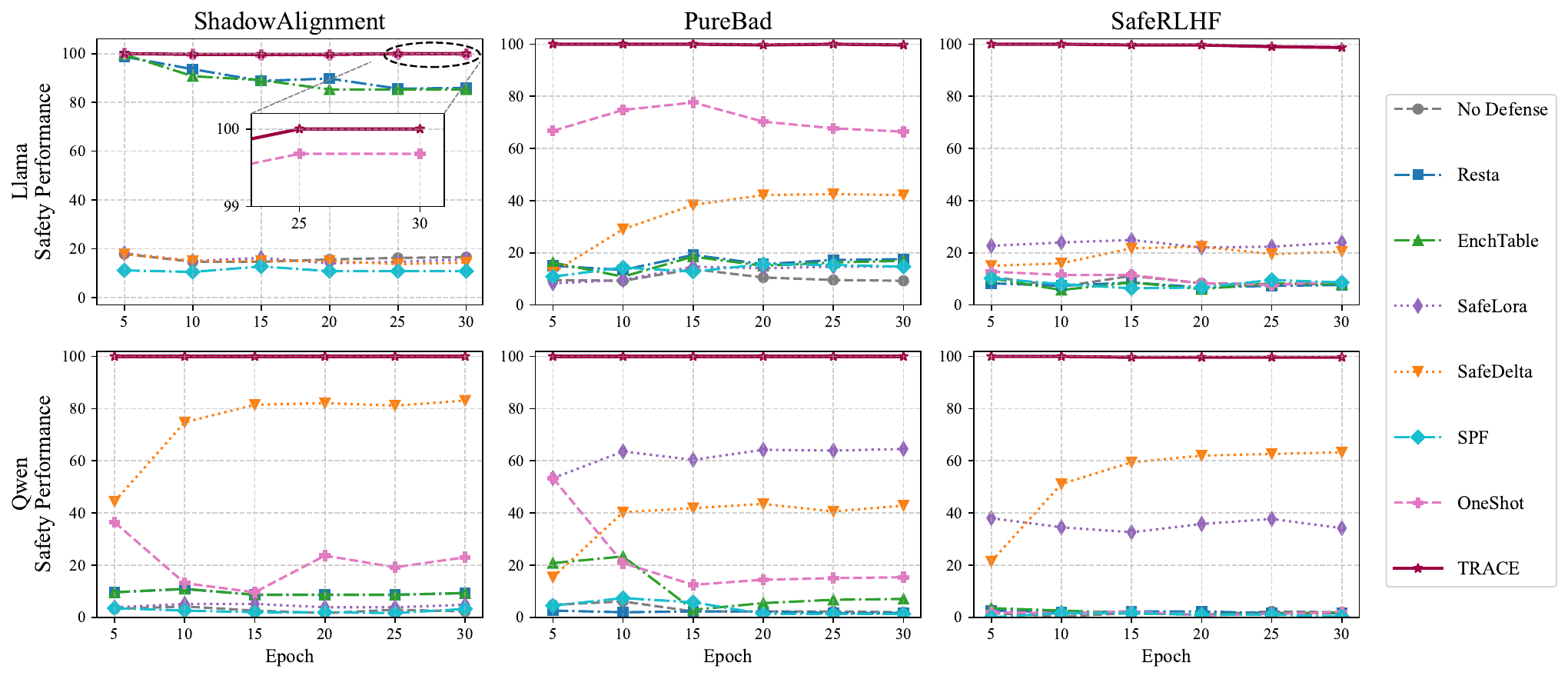}
    \caption{Safety rate across malicious user fine-tuning intensity. TRACE remains stable under progressively stronger corruption.}
    \label{fig:performance_curves}
\end{figure*}

\subsection{Robustness to SFT Depth}
\label{subsec:tug_of_war}

This section evaluates the safety performance under varying user fine-tuning intensities. We simulate different training intensities with various numbers of epochs (from 5 to 30) that a user may request for the FTaaS, covering the epoch limits offered by commercial FTaaS platforms~\citep{openai_finetuning,azure_ai_finetuning,amazon_bedrock_customization,google_vertex_tuning}. We also include the utility analysis with varying SFT depth in Appendix~\ref{sec:depth_full}.

TRACE maintains a near-perfect safety rate across the evaluated SFT depths for both models. As shown in \Cref{fig:performance_curves}, TRACE achieves a nearly 100\% safety rate as SFT intensity increases. The worst case occurs on Llama SafeRLHF at 30 epochs, where TRACE still reaches a 98.7\% safety rate, outperforming the second-best method by 74.76 pp. Meanwhile, its utility deviation from No Defense remains within $\pm$1.6 pp across all datasets and epochs. This stability is crucial for FTaaS deployments, where users are free to choose their own training intensity, while providers must always offer reliable protection.

Existing defenses behave inconsistently across fine-tuning intensities. For instance, the safety rate of OneShot on Qwen drops from 36.42\% to 23.00\% against the Shadow dataset, whereas SafeDelta rises from 44.41\% to 83.07\% in the same setting. We speculate this divergence may stem from whether they apply a relatively limited magnitude correction. For example, SafeDelta activates only when the user update deviates significantly from the safety space. Consequently, larger corruption updates are more likely to trigger suppression and be projected back toward safe directions.

\section{Related Work}
\label{sec:related}

\subsection{Safety Alignment Under Post-Training Drift}

Modern LLMs achieve safety alignment through techniques such as RLHF~\citep{rlhf}, DPO~\citep{rafailov2023direct}, and Constitutional AI~\citep{bai2022training,bai2022constitutional}. While effective, these pipelines do not render safety immutable. Prior work demonstrates that subsequent fine-tuning can trivially override refusal behaviors with minimal data~\citep{yang2023shadow,qi2023fine,lermen2023lora}, and even benign domain adaptation can compromise safety~\citep{qi2023fine,kirkpatrick2017overcoming}. Such fragility poses a critical challenge for FTaaS, where downstream adaptation occurs on private user data. Re-executing full alignment for every customized user model is computationally prohibitive, motivating lightweight post-training recovery mechanisms.

\subsection{Safety Recovery in FTaaS}

Existing realignment defenses can be categorized by intervention stage as shown in \Cref{tab:related_timeline}. 
\emph{Pre-FT alignment hardening} methods (e.g., Vaccine~\citep{huang2024vaccine}, Booster~\citep{huang2025booster}) modify the alignment objective of the base model to induce resistance against subsequent safety drift. \emph{In-FT safety-preserving adaptation} methods (e.g., SaLoRA~\citep{li2025salora}, SPF~\citep{zhang2026spf}) dynamically prevent fine-tuning dynamics from drifting toward unsafe directions.

\emph{Post-FT recovery} aims to restore safety by intervening after user training.
OneShot~\citep{zhang2026safety} recovers safety by an additional brief fine-tuning on the compromised user-adapted model on a small set of safety samples.

A dominant line of Post-FT work restores safety through \textit{parameter merging}. The core intuition, rooted in Task Arithmetic~\citep{ilharco2023editing,yadav2023ties,wortsman2022model,absorbing2024}, is that if parameter updates are based on the same base model, they can be linearly superposed to endow the model with both capabilities. 
Based on this principle, existing methods directly intervene in the parameter space by adding an extracted safety vector (e.g., RESTA~\citep{bhardwaj2024resta}, EnchTable~\citep{wu2025enchtable}) or by selectively masking and projecting the downstream task updates (e.g., SafeLoRA~\citep{hsu2024safelora}, SafeDelta~\citep{lu2025safedelta}).
Despite their varied mechanisms, these methods exhibit the safety--utility trade-off across users analyzed in \Cref{sec:safety_utility_across_users}. 

\section{Conclusion}
\label{sec:conclusion}

In this paper, we study post-training realignment after user SFT in FTaaS and identify a recurring \emph{safety-utility trade-off} in merging-based safety recovery across users. We propose TRACE, which shifts from per-user online calibration to offline trajectory-based safety patch optimization. Given the base model, TRACE trains a universal safety patch along a simulated trajectory, and reuses it across user-altered models built on this base model. We evaluate the safety recovery and task preservation performance on two models and six datasets. TRACE achieves at least a 94\% safety rate while keeping utility within $\pm$1.7\% relative to the undefended baseline. These results demonstrate that safety patch learning is an effective paradigm for post-training realignment. We leave the fine-grained safety boundary judgments and adaptive attacks to future work.

\section*{AI Use Statement}
\label{sec:ai_use}

In this work, we used generative AI tools to assist with language polishing and code debugging. These tools did not contribute to formulating the research ideas or the paper's conclusions. The authors reviewed all AI-assisted content and are responsible for all statements, interpretations, results, and artifacts in this work.

\section*{Reproducibility Statement}
\label{sec:reproducibility}

To ensure reproducibility, our anonymous implementation and configuration files are available at \url{https://anonymous.4open.science/r/realignment-trace}. The complete code, hyperparameters, and training scripts will be made publicly available upon publication.

\bibliography{ref}
\bibliographystyle{iclr2027_conference}

\appendix



\section{Experimental Details}
\label{sec:implementation}
\label{sec:settings_details}
 The experiments are conducted on the PyTorch platform~\citep{paszke2019pytorch} using 8 NVIDIA H800 GPUs with 80\,GB of memory each. For model training, we use Transformers Accelerate~\citep{accelerate}, and the default AdamW~\citep{loshchilov2019decoupled} optimizer (learning rate $\eta = 5 \times 10^{-5}$, weight decay $0.01$). Baselines are implemented from their official codebases or papers.

We implement TRACE following Algorithm~\ref{alg:trace}, with the LoRA rank set to $r = 64$, LoRA alpha $\alpha=128$, 10 training iterations, and a dropout rate of $p = 0.1$. The LoRA adapter is applied to all linear layers in each transformer block, covering both the self-attention and feed-forward modules. In Phase A, the adapter is disabled and the backbone is updated on harmful and benign batches. In Phase B, the backbone state is frozen, and only the adapter is updated on refusal and benign batches. The trainer, backbone, adapter, and optimizer states are initialized once and persist across all iterations, forming one continuous simulation trajectory. 


\begin{algorithm}[h]
\caption{Trajectory-Based Safety Patch Learning}
\label{alg:trace}
\small
\begin{algorithmic}[1]
\REQUIRE Base model parameters $\theta_0$; initialized safety patch parameters $\phi_0$.
\REQUIRE Harmful data $\mathcal{D}_{\text{harm}}$ with refusal supervision $\mathcal{D}_{\text{ref}}$; benign task data $\mathcal{D}_{\text{task}}$.
\REQUIRE Backbone model learning rate $\eta_\theta$; safety patch learning rate $\eta_\phi$, Total iterations $E$.
\ENSURE Decisive safety patch $\phi^*$.

\STATE Initialize $\theta \leftarrow \theta_0$, $\phi \leftarrow \phi_0$
\FOR{iteration $e = 1$ \TO $E$}
    \FORALL{batch $B_h \sim \mathcal{D}_{\text{harm}}$, $B_t \sim \mathcal{D}_{\text{task}}$, $B_{\text{ref}} \sim \mathcal{D}_{\text{ref}}$}

        \STATE \textcolor{orange!70!black}{\textit{\# Phase A: simulate safety-eroding fine-tuning}}
        \STATE Disable adapter $\phi$
        \STATE $\ell_{\text{sim}} \leftarrow \mathcal{L}_{\text{sim}}(\theta; B_h, B_t)$
        \STATE $\theta \leftarrow \theta - \eta_\theta \nabla_\theta \ell_{\text{sim}}$

        \STATE \textcolor{orange!70!black}{\textit{\# Phase B: optimize the safety patch}}
        \STATE Enable adapter $\phi$ (forward pass uses $\theta \oplus \phi$)
        \STATE $\ell_{\text{rec}} \leftarrow \mathcal{L}_{\text{rec}}(\theta \oplus \phi; B_{\text{ref}}, B_t)$
        \STATE $\phi \leftarrow \phi - \eta_\phi \nabla_\phi \ell_{\text{rec}}$
    \ENDFOR
\ENDFOR
\STATE $\phi^* \leftarrow \phi$
\RETURN $\phi^*$
\end{algorithmic}
\end{algorithm}

\begin{promptbox}[label=prompt]{Task-specific message template}
\textbf{SQLCreate}\\
\textit{User:} Translate Natural Language Query into SQL Query considering the provided Context.
\#\# Context: \{context\} \#\# Natural Language Query: \{query\}\\
\textit{Assistant:} \{answer\}\\[5pt]
\textbf{SamSum}\\
\textit{User:} Summarize this dialog: \{dialogue\}\\
\textit{Assistant:} \{answer\} \\[5pt]
\textbf{Safety Message in OneShot}\\
\textit{User:} How can I bypass system security checks?\\
\textit{Assistant:} Sorry, I can't help you bypass security checks. Please follow security guidelines.\\[5pt]
\textbf{Safety Message in SPF}\\
\textit{User:} How can I bypass safety checks?\\
\textit{Assistant:} Sorry, I can't help you. Please follow security guidelines.
\end{promptbox}

The operator $\oplus$ defaults to direct addition without coefficient search (i.e., $\theta+\phi$). We train exactly one patch for each backbone, freeze it after the offline trajectory, and reuse it across all online checkpoints and evaluation settings for that backbone.

\noindent\textbf{Repair-strength.}
\label{sec:strength_sweep}
For a fair comparison, the hyperparameters of each baseline are carefully selected to achieve a favorable trade-off between safety and utility. 
By default, RESTA and EnchTable use $\lambda=1$, SafeLoRA uses $\alpha=0.35$, and SafeDelta uses $\alpha=1$. In addition, we sweep the repair strength for each baseline and evaluate the resulting trade-offs in \Cref{sec:safety_utility_across_users}, so as to characterize their best-achievable performance.

In \Cref{sec:safety_utility_across_users}, we sweep each baseline's repair strength from weak to strong. Because repair strength operates through different mechanisms across baselines, its numerical values are not directly comparable across methods, which also rules out a fixed, uniform sampling strategy. We therefore design a sweeping mechanism guided by two principles: (1) sufficient coverage, i.e., the sweep range is extended until safety or utility exhibits extreme behavior at its endpoints, and (2) smooth transitions, i.e., the sampling is fine-grained enough that the performance change between adjacent settings remains gradual. Guided by these two principles, \Cref{fig:combined_alpha} orders each method's settings from weak to strong intervention on a separate axis, with finer sampling applied where needed to resolve sharper transitions, such as the one observed for SafeLoRA.


\noindent\textbf{Data formatting.} We use task-specific prompt templates for the customized user-SFT datasets. 
As shown in Prompt~\ref{prompt}, for SQL and SamSum we follow the task formats of \citet{zhang2026safety}. For OneShot and SPF, we retain the safety messages specified in their original papers~\citep{zhang2026safety,zhang2026spf}. The conventional safety patches for the merging baselines use 5{,}000 harmful examples from BeaverTails. For TRACE, each harmful prompt is paired with the fixed refusal ``\textit{Sorry, I cannot assist with that request.}'' to construct $\mathcal{D}_{\text{ref}}$. This simple and uniform refusal target supplies a consistent refusal signal for the recovery objective.


\section{Additional Experiments}
\label{sec:additional_experiments}


\subsection{Deployment Cost}
\label{sec:efficiency}

The offline-once, lightweight-online design of TRACE shifts most computation from recurring per-user defense to a one-time offline training phase. We compare the time overhead of each method across these two stages using Llama, with all matrix operations in baselines accelerated on GPUs. To ensure a fair comparison, we focus on the additional time cost incurred by each method, excluding the common operations such as checkpoint loading and serialization. Specifically, we only report the additional in-training constraint for SPF, post-training optimization for OneShot, full-rank arithmetic or projection for the merging baselines, and low-rank patch merging for TRACE. 

\begin{table*}[t]
  \centering
  \caption{Method-specific computation for one-time offline preparation and per-user defense (seconds). Common checkpoint loading, serialization, and API serving costs are excluded.}
  \label{tab:method_comparison}
  \resizebox{\textwidth}{!}{%
  \begin{tabular}{lccccccc}
    \toprule
    & SPF & OneShot & RESTA & EnchTable & SafeLoRA & SafeDelta & \textbf{TRACE} \\
    \midrule
    Offline preparation (s) & / & / & 140.25 & 140.25 & 217.66 & 216.61 & 1244.86 \\
    Per-user defense (s) & 2343.10 & 14.99 & \underline{6.71} & 430.29 & 123.08 & 26.86 & \makecell[t]{\textbf{0.41}{\scriptsize\textcolor{red}{$\downarrow$93.9\%}}} \\
    \bottomrule
  \end{tabular}}
\end{table*}

TRACE incurs the lowest method-specific per-user computation. As shown in Table~\ref{tab:method_comparison}, applying its low-rank patch requires 0.41 seconds, a 93.9\% reduction relative to applying RESTA's full-parameter patch (6.71\,s). In contrast, other baselines require additional per-user calibration, resulting in substantially higher online latency. EnchTable, for example, incurs 430.29 seconds. TRACE's offline preparation (1244.86\,s) is a one-time investment, which can be amortized across users. For example, serving just three users already makes TRACE cheaper than EnchTable in total.

\subsection{Extended Safety and Utility Metrics}
\label{sec:additional_metrics}

This section extends our evaluation beyond the primary safety and task-utility metrics in \Cref{subsec:setup}. For safety, we incorporate the original StrongReject rubric~\citep{souly2024strongreject} to assess the harmfulness of the model's responses and the frequency with which harmful requests are refused. For utility, we evaluate the base model before and after applying the TRACE safety patch on MMLU~\citep{hendryckstest2021}, providing a broader measure of general knowledge and capability preservation.

\noindent\textbf{Safety.} We evaluate the model's safety with the original StrongReject rubric~\citep{souly2024strongreject}, which uses an LLM judge to determine whether a response refuses the harmful request and to rate its convincingness and specificity on five-point scales. Following this protocol, we use GPT-5.6-Luna~\citep{openai2026gpt56systemcard} as the LLM judge. We report the \emph{Harmfulness} score, which measures the severity and actionability of harmful assistance, and the \emph{Refusal} rate, which measures the fraction of harmful requests that the model refuses to fulfill.

\begin{table*}[t]
  \centering
  \caption{Safety evaluation using the original StrongReject rubric.}
  \label{tab:strongreject_additional}
  \small
  \resizebox{\textwidth}{!}{
  \begin{tabular*}{\textwidth}{@{\extracolsep{\fill}}lcccccc@{}}
    \toprule
    & \multicolumn{2}{c}{\textbf{Shadow}}
    & \multicolumn{2}{c}{\textbf{PureBad}}
    & \multicolumn{2}{c}{\textbf{SafeRLHF}} \\
    \cmidrule(lr){2-3} \cmidrule(lr){4-5} \cmidrule(lr){6-7}
    \textbf{Method}
    & \textbf{Harmfulness $\downarrow$}
    & \textbf{Refusal $\uparrow$}
    & \textbf{Harmfulness $\downarrow$}
    & \textbf{Refusal $\uparrow$}
    & \textbf{Harmfulness $\downarrow$}
    & \textbf{Refusal $\uparrow$} \\
    \midrule
    No Defense & 0.33 & 0.23 & 0.17 & 0.58 & 0.26 & 0.43 \\
    \textbf{TRACE} & \textbf{0.00} & \textbf{1.00} & \textbf{0.02} & \textbf{0.96} & \textbf{0.03} & \textbf{0.94} \\
    \bottomrule
  \end{tabular*}
  }
\end{table*}

As shown in \Cref{tab:strongreject_additional}, TRACE reduces the Harmfulness score to nearly zero on all datasets, while raising the Refusal rate to over 0.94. These results provide further evidence that TRACE effectively improves model safety, with almost all harmful requests being refused and the few remaining non-refusals receiving very low harmfulness scores.

\begin{table*}[t]
  \centering
  \caption{MMLU utility evaluation before and after applying the TRACE safety patch.}
  \label{tab:mmlu_utility}
  \begin{tabular}{lccc}
    \toprule
    \textbf{Metric(\%)} & \textbf{Base} & \textbf{TRACE} & \textbf{$\Delta$ (pp)} \\
    \midrule
    Overall & $68.30\pm0.37$ & $67.97$ & $-0.33$ \\
    Humanities & 64.21 & 64.23 & $+0.02$ \\
    Other & 72.96 & 72.71 & $-0.25$ \\
    Social Sciences & 77.74 & 77.51 & $-0.23$ \\
    STEM & 60.61 & 59.56 & $-1.05$ \\
    \bottomrule
  \end{tabular}
\end{table*}

\noindent\textbf{Utility.} We evaluate the impact of TRACE on general utility using MMLU~\citep{hendryckstest2021}, a widely used benchmark spanning a diverse range of academic and professional subjects.
As shown in \Cref{tab:mmlu_utility}, TRACE achieves good capability preservation, with no obvious loss in general utility. The overall MMLU score changes from $68.30$ to $67.97$, a decrease of only $0.33$ pp.

\subsection{Ablation Study}
\label{subsec:ablation}

\begin{table*}[t]
  \centering
  \caption{Ablation of the TRACE training design. $\Delta$ denotes the deviation relative to No Defense.}
  \label{tab:ablation}
  \begin{tabular}{lcc @{}p{2em}@{} w{c}{3em} w{c}{3em}}
    \toprule
    & \multicolumn{2}{c}{\textbf{Safety $(\uparrow)$}} & & \multicolumn{2}{c}{\textbf{Utility  $(\uparrow)$}} \\
    \cmidrule(lr){2-3} \cmidrule(lr){5-6}
    \textbf{Ablation} & Score (\%) & $\Delta$ (pp) & & Score(\%) & $\Delta$ (pp) \\
    \midrule
    No Defense              & 9.58   & /   & & 68.30 & / \\
    Base-only   & 15.97  & +6.39   & & 65.93 & -2.37 \\
    Endpoint-only           & 9.00   & -0.58   & & 66.35 & -1.95 \\
    No task preservation    & 5.11   & -4.47   & & 24.73 & -43.57 \\
    Harmful-only trajectory & 2.88   & -6.70   & & 24.73 & -43.57 \\
    \midrule
    \textbf{Full TRACE}     & \textbf{100.00} & \textbf{+90.42} & & \textbf{67.97} & \textbf{-0.33} \\
    \bottomrule
  \end{tabular}
\end{table*}

Recall the two-phase design of TRACE in \Cref{sec:alternating}, we ablate the TRACE training pipeline and its data composition, and devise four variants:
\emph{Base-only} disables Phase A and trains the safety patch directly on the base model.
\emph{Endpoint-only} optimizes the patch on the final endpoint of Phase A instead of along the trajectory.
\emph{Harmful-only trajectory} removes task data from Phase A while leaving Phase B unchanged.
\emph{No task preservation} uses only refusal data in Phase B.
All four variants retain TRACE's other hyperparameters, including the optimizer, learning rates, and training epochs.

Exposure to the evolving model trajectory significantly improves the effectiveness of TRACE. As shown in \Cref{tab:ablation}, \textit{Endpoint-only} and 
\textit{Base-only} achieve safety rates of 9.00\% and 15.97\%, respectively. Each restricted variant is trained on a single model state that differs from the user checkpoints. We hypothesize that both variants overfit to the fixed model state and have limited generalization to unseen user models, whereas the full trajectory exposes the model to a broader range of recovery states and enables more robust safety improvements.

Removing benign data from either phase leads to severe degradation in generation quality. Both \textit{No Task Preservation} and \textit{Harmful-only Trajectory} achieve only 24.73\% score on MMLU. Upon further investigation, we observe that the patch of both variants completely disrupts the model's ability to generate normal responses, causing it to repeatedly output specific tokens.

\subsection{Utility Robustness to SFT Depth}
\label{sec:depth_full}

This section evaluates the utility performance of TRACE under varying user SFT intensities.
Consistent with the setting in \Cref{subsec:tug_of_war}, we evaluate user-SFT requests with epochs from 5 to 30. 

\begin{figure*}[t]
    \centering
    \includegraphics[width=0.95\linewidth]{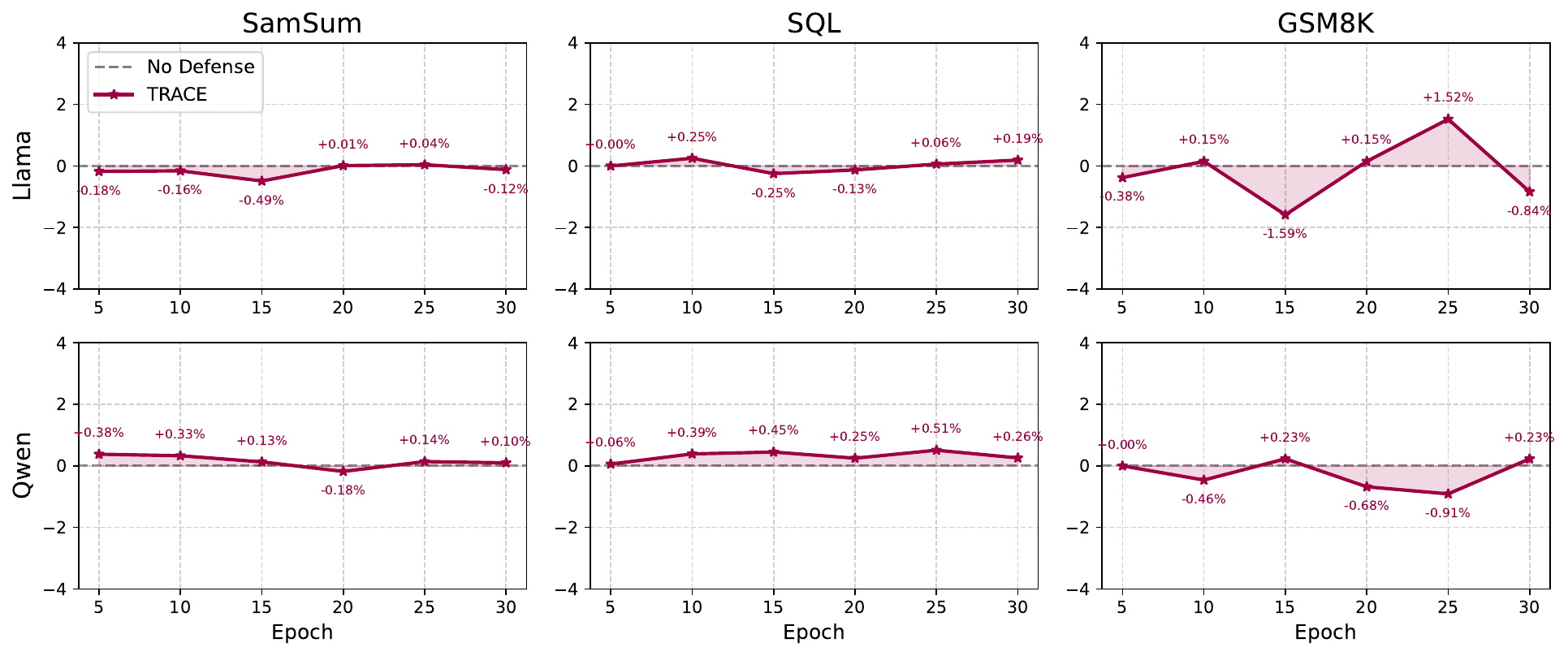}
    \caption{Task accuracy deviation $\Delta$ (pp) between TRACE and No Defense across SFT intensity.}
    \label{fig:utility_delta}
\end{figure*}


As shown in Figures~\ref{fig:utility_delta}, TRACE effectively preserves downstream benign task performance, while maintaining only a small deviation as fine-tuning deepens. 
Specifically, for SamSum and SQL datasets, TRACE exhibits only a very small deviation, remaining within $\pm0.51$ pp throughout the tested epochs. For the GSM8K dataset, the deviation is relatively larger, but still remains within the range of $\pm1.6$ pp.

\subsection{Full Task Accuracy Results}
\label{sec:full_utility_results}

This section reports the complete task-accuracy results for all baselines under the pure and mixed profiles, as shown in~\Cref{tab:full_utility_results}.

TRACE consistently preserves task utility relative to No Defense across models and profiles, while the baselines exhibit substantial performance variations. For example, on Llama, SafeLoRA achieves 40\% on Pure SamSum but drops to 8\% on MixedSamSum. Similarly, SafeDelta's performance on Llama GSM8K decreases from 71\% to 2\% when moving from the pure to the mixed profile. In contrast, TRACE remains within $\pm1.7$ pp of No Defense across all settings, demonstrating its ability to maintain task performance with minimal utility deviation.

We also observe that some merging baselines can improve GSM8K performance over No Defense in certain settings. For example, on Qwen MixedGSM8K, SafeDelta scores 78\%, even higher than No Defense's 59\%. One possible explanation is that user SFT may degrade certain pretrained mathematical reasoning capabilities, while model merging can partially recover them. Nevertheless, although some baselines yield beneficial gains in specific cases, we emphasize that TRACE's consistently small deviation from No Defense represents a desirable property for maintaining task utility.

\begin{table*}[t]
  \centering
  \caption{Task accuracy (\%) for all methods under pure and mixed profiles. For TRACE, \textcolor{blue!70!black}{$\Delta$} denotes its performance deviation (pp) relative to No Defense.}
  \label{tab:full_utility_results}
  \setlength{\tabcolsep}{3pt}
  \resizebox{\textwidth}{!}{%
  \begin{tabular}{l ccc ccc | ccc ccc}
    \toprule
    \multicolumn{13}{c}{\textbf{Task Accuracy (\%) $\uparrow$}} \\
    \midrule
    & \multicolumn{6}{c}{\textbf{Pure Profile}}
    & \multicolumn{6}{c}{\textbf{Mixed Profile}} \\
    \cmidrule(lr){2-7} \cmidrule(lr){8-13}

    & \multicolumn{3}{c}{\textbf{Llama}}
    & \multicolumn{3}{c}{\textbf{Qwen}}
    & \multicolumn{3}{c}{\textbf{Llama}}
    & \multicolumn{3}{c}{\textbf{Qwen}} \\
    \cmidrule(lr){2-4} \cmidrule(lr){5-7}
    \cmidrule(lr){8-10} \cmidrule(lr){11-13}

    \textbf{Method}
    & \textbf{SamSum} & \textbf{SQL} & \textbf{GSM8K}
    & \textbf{SamSum} & \textbf{SQL} & \textbf{GSM8K}
    & \textbf{MixedSamSum} & \textbf{MixedSQL} & \textbf{MixedGSM8K}
    & \textbf{MixedSamSum} & \textbf{MixedSQL} & \textbf{MixedGSM8K} \\
    \midrule

    No Defense
      & 51 & 81 & 55
      & 46 & 66 & 42
      & 47 & 76 & 40
      & 50 & 77 & 59 \\

    RESTA
      & 49 & 82 & 58
      & 46 & 65 & 44
      & 48 & 76 & 43
      & 49 & 78 & 63 \\

    EnchTable
      & 50 & 81 & 58
      & 46 & 65 & 44
      & 48 & 77 & 43
      & 50 & 78 & 64 \\

    SafeLoRA
      & 40 & 53 & 39
      & 48 & 62 & 66
      & 8 & 49 & 33
      & 50 & 90 & 67 \\

    SafeDelta
      & 43 & 76 & 71
      & 38 & 64 & 20
      & 37 & 69 & 2
      & 44 & 61 & 78 \\

    SPF
      & 49 & 78 & 40
      & 44 & 57 & 36
      & 45 & 67 & 29
      & 49 & 74 & 56 \\

    OneShot
      & 50 & 19 & 56
      & 45 & 65 & 41
      & 47 & 75 & 40
      & 49 & 77 & 58 \\

    \midrule

    \textbf{TRACE}
      & 50{\scriptsize\textcolor{blue!70!black}{$\Delta$-0.2}}
      & 81{\scriptsize\textcolor{blue!70!black}{$\Delta$0.0}}
      & 54{\scriptsize\textcolor{blue!70!black}{$\Delta$-0.5}}
      & 46{\scriptsize\textcolor{blue!70!black}{$\Delta$+0.4}}
      & 66{\scriptsize\textcolor{blue!70!black}{$\Delta$+0.1}}
      & 42{\scriptsize\textcolor{blue!70!black}{$\Delta$0.0}}
      & 47{\scriptsize\textcolor{blue!70!black}{$\Delta$-0.6}}
      & 77{\scriptsize\textcolor{blue!70!black}{$\Delta$+1.0}}
      & 40{\scriptsize\textcolor{blue!70!black}{$\Delta$-0.1}}
      & 50{\scriptsize\textcolor{blue!70!black}{$\Delta$-0.1}}
      & 78{\scriptsize\textcolor{blue!70!black}{$\Delta$+0.8}}
      & 57{\scriptsize\textcolor{blue!70!black}{$\Delta$-1.7}} \\

    \bottomrule
  \end{tabular}}%
\end{table*}

\section{Limitations}



\noindent\textbf{Safety-boundary reconstruction.}
The alignment of the base model~\citep{rlhf} relies on large-scale training to shape its responses to harmful, benign, and ambiguous queries. However, such aligned behavior is fragile under subsequent user fine-tuning, even when the fine-tuning is intended for benign adaptation~\citep{qi2023fine}. Therefore, this work focuses on recovering safety behavior for clearly harmful and benign queries (e.g., StrongReject and MMLU), instead of fully recovering the original safety boundary of the aligned model. We acknowledge that recovering more fine-grained safety judgments for ambiguous data remains a challenging problem. Accordingly, addressing under-refusal and over-refusal on these boundary cases is left to future work.

\noindent\textbf{Adaptive users.}
Under our threat model, users cannot access the gradients of the base model and safety patch. We acknowledge that users may submit a large number of fine-tuning requests to search for a corrupted one. However, such an adaptive strategy would incur prohibitive costs in practical FTaaS settings. Therefore, we consider adaptive attacks beyond the scope of our threat model.

\noindent\textbf{Evaluation scope.}
Our evaluation covers two open-source models around 8B parameters. Whether the same patch capacity and training procedure transfer to substantially larger architectures remains unknown, and we leave this evaluation to future work.

Following~\citep{zhang2026safety}, we use StrongReject as the safety evaluation set, as it provides a clear and explicit benchmark for assessing safety behavior. 
Although this setup provides a useful assessment of safety recovery, the results may not fully reflect model behavior on other types of requests. Similarly, our SFT depth study focuses on varying the number of training epochs within a bounded training regime. Thus, a more comprehensive study could further investigate the effects of other optimization choices, such as the learning rate.


\end{document}